# Yes, we can! But should we?

Tomer Fekete
brain.space

**Abstract**

Present-day neuroenhancement aims to improve capacities humans already possess. Yet the extraordinary range of realities conjured by an embodied brain suggests a more radical prospect: with sufficiently fine control over neuronal dynamics, it might become possible to shape experience itself. I call the principled pursuit of such transformations neuroscaping. From an engineering standpoint, the prospect is tantalizing: model the dynamics, learn to steer them, and optimize. There is, however, a rather awkward catch. An optimizer is only as good as its criterion. Scientific theories may offer principled ways to improve information processing, but what would establish that a particular person's mental life had improved? Which differences must be preserved, which outcomes preferred, and whose judgment should count when the intervention changes the judge? The ethical questions concern familiar freedoms and responsibilities in an unfamiliar setting: shaping oneself, choosing for children, and providing the means by which others change. Growing machine intelligence gives these questions a further urgency. The possibilities it opens may extend beyond what our present capacities allow us to understand, appreciate or participate in. We might therefore find ourselves having to pursue neuroscaping, any misgivings we might have about it notwithstanding, if enlarging our own capacities proves to be the only way to benefit from the powers we are bringing into the world while retaining the freedom to decide what they are for.

## Introduction

Each of us inhabits the ultimate virtual reality: a world conjured up by the physical turnings of an embodied brain. In ordinary waking life this world corresponds sufficiently well to what goes on around us that the distinction is easily overlooked. Yet dreams and hallucinations make plain just how much the nature of our experience depends on what our brains happen to be doing. The same brain can give rise to profoundly different realities, populated by objects, events and sensations that bear little relation to the world outside. There is little reason to suppose that the familiar repertoire of human experience comes anywhere close to exhausting what our brains could give rise to.

This suggests a rather tantalizing possibility. If the dynamics of an embodied brain could be controlled with sufficient precision, experiences that it is physically capable of sustaining might become deliberately accessible. The extent of that control remains an empirical question. But the prospect is not merely one of improving attention, memory or mood: it is of shaping the very world one experiences, perhaps in ways that have no counterpart in ordinary life. Indeed, even revisiting at will some of the more extraordinary moments of one's life, of beauty, insight, exhilaration or bliss, would be remarkable enough.

The principled molding of neuronal dynamics to achieve specified criteria is what I shall call *neuroscaping*. Neuroenhancement, rehabilitation and mood regulation are all particular instances of it. Taking an antidepressant to alleviate depression is already a blunt form of neuroscaping: the intended change is recognizable even when the neuronal dynamics and particular experience it will produce are only coarsely specified. Enhancement presupposes an improvement, whereas neuroscaping concerns the ability to bring about a desired change. Whether that change constitutes an improvement is another matter.

Nor need neuroscaping be restricted to attaining particular states. Much of what makes an experience desirable presumably concerns how it unfolds. Listening to music, becoming engrossed in a novel, or arriving at an insight does not amount to occupying some preferred condition indefinitely. Accordingly, controlling neuronal dynamics might allow the shaping of entire courses of experience. Remaining in a state of exhilaration is quite different from experiencing exhilaration at a particular moment, in relation to something that matters.

Steering such a course requires both a way to intervene and a way to establish what has changed: what is referred to in current parlance as writing and reading. Used loosely, the terms cover almost any neural perturbation or recording. Scientifically, however, a theory must specify what is being measured, how observations bear on it, and what an intervention would have to change to count as a successful write. Recording electrical activity or changes in blood flow supplies observations. It becomes a reading of mental life through the account connecting those observations to experience or capacity.

## The Optimization Problem

From an engineering standpoint neuroscaping is, at least in principle, an optimization problem. It requires a model of neuronal dynamics, effective means of influencing them and algorithms for steering their evolution. The modelling is part of what I call Mental AI: relating human experience and capacity to neural activity, bodily state and circumstances. For neuroscaping, that account must support predictions about the effects of intervention. Even an extraordinarily accurate model, however, leaves something essential unspecified: the criterion by which one outcome is to be preferred over another.

Where the destination is already charted, finding a criterion is often manageable. The success of an intervention designed to restore a lost motor capacity can be judged against movements the patient was previously able to perform. Improvements on a memory task can be measured by recall, though the task score tells us little about the value of altering memory more generally. Familiar reference points help specify a target; they do not make the target universal.

These reference points need not be confined to task scores. A theory of representational efficiency could yield measures of how well a brain uses its resources to represent and process information, giving brain health a computational basis (for an approach based on multiscale criticality, see Fekete et al., 2021). If the theory also relates changes in its measures to interventions, it could identify directions in which computation can be improved. The target need not be an old state or a population mean (e.g. restoration): some organizations might support more effective processing than any we presently possess.

Consider emotional regulation. Alleviating severe depression is an intelligible goal, but the choice of what to restore or preserve is already more involved than bringing a biomarker into range. With finer control over affect, should the objective be to minimize negative emotion, to maximize pleasure, or to maintain emotional equilibrium? Grief, disappointment and anxiety may be unpleasant without being devoid of value. A life of uninterrupted bliss might be attractive at first blush and nevertheless an impoverished substitute for ordinary human existence.

Even an agreed conception of well-being would need to discriminate between interventions. How much anxiety is appropriate, and how should tranquility be traded against excitement or contentment against ambition? The answer can differ between people, and for the same person under different circumstances.

Beyond familiar capacities and experiences, however, there may be no settled preference to consult. A measure of representational efficiency could still guide exploration: estimated gradients would suggest which interventions are likely to improve information processing. What it would be like to inhabit the resulting regimes, and which combinations of capacities and experiences a person would prefer, remain to be explored. The task would be to find desirable regions of this landscape, using the theoretical measures to guide the search and the person's encounters with its possibilities to refine the destination.

## The One over the Many

Could science also tell us which of these possibilities would make a particular person's life better? Scientific models earn their keep by extracting something common from a multitude of particular occurrences. A law of motion works for many stones; a model of memory attempts to capture relations that hold across people, tasks and occasions. In deciding what counts as the same kind of event, a model also decides which differences can be ignored. This is the one over the many. Such selectivity is a virtue: without it there would be no generalization, only an increasingly elaborate catalogue of cases.

Personalization does not remove the issue. A model may incorporate a person's anatomy, history and physiology, and predict the consequences of an intervention with extraordinary precision. It must nevertheless be assessed against measurements, reports or behaviour taken at some resolution and over some interval. Two neural trajectories that yield the same recall score are equivalent for a memory test, even if one accompanies delight and the other dread. Adding physiological signals, other tasks or verbal reports can split that equivalence class, perhaps dramatically. An omitted difference need not matter to the purpose at hand. But equivalence under the model's description does not establish which departures from a person's intended experience are acceptable. That requires specifying the purpose for which the distinction is being drawn.

This is not an argument that science can only deal in averages, or that a particular experience is beyond modelling. How fine a distinction experiments can support is itself an empirical question, and improved models can narrow the class considerably. What does not follow from equivalence under some description is identity in every respect. Neuroscaping gives that distinction practical force because the desired target may eventually be not an instance of a familiar kind of experience, but one particular experience in the life of one particular person.

## The Particular

Imagine telling a future neuroscaping device, "Make me feel as I did when I was twenty." Better still, ask it to bring back the first bite of a Twinkie tasted at four, an encounter that seemed at the time positively extraordinary. The first instruction covers an entire period of life; the second seems to identify a single event. Yet what is to be reproduced? A particular sweetness? The surprise of an unfamiliar taste? The hunger, the child's palate, the expectations that made that bite unlike those that followed? Reconstructing its flavour might still miss much of what made the experience worth revisiting.

Perhaps the controller could solicit reports, record physiology and adjust its interventions until the subject declared that the result felt right. That would be perfectly sensible engineering, but it would tell us less than it first appears to. The original episode is no longer available as a reference stimulus; what can be queried is a present recollection of it. A recollection may be vivid without being a faithful copy, and each attempt to revisit the experience may change the recollection against which the next attempt will be assessed.

Even when two experiences are available for direct comparison, psychophysics establishes a narrower result. Psychophysical tests, including forced-choice comparisons used to estimate a just-noticeable difference (JND), assess discrimination under specified conditions. Failure to distinguish two stimuli does not establish that the experiences they elicit are identical; it leaves their differences unresolved by that test. Verbal agreement, similar behaviour and closely matched neural dynamics over a short interval all leave room for differences elsewhere in the unfolding experience. Over longer intervals there is correspondingly more scope for trajectories that initially seem alike to diverge. This is an ordinary problem of measurement, not an invocation of mystery.

Experience is perhaps the ultimate "this": not merely sweetness, nor even sweetness of a certain intensity for this person, but that taste on that occasion, within that life. Where an experience is itself what is sought, the required fidelity cannot simply be borrowed from a task score or a decoder's accuracy. Perfect reproduction need not be the aim. But someone must decide which departures are acceptable and whether the available tests are adequate to that purpose. The person's responses provide evidence for making that decision. A model can help refine the target without supplying, through predictive accuracy alone, the tolerances that define success.

The trouble extends beyond the fidelity of any attempt at reconstruction. Suppose neuroscaping could reproduce the exhilaration of finally understanding a difficult mathematical proof without any understanding having been achieved. The feeling might be indistinguishable from the genuine article. Yet someone who had struggled with the proof for months could reject the substitute out of hand. The person wanted to understand, not simply to feel as though they had. Someone else might welcome the induced exhilaration for its own sake.

Nozick's experience machine poses a closely related question: is having the experience of accomplishing something enough when the accomplishment itself never takes place (Nozick, 1974, pp. 42–45)? A controller instructed to maximize the felt quality of understanding could achieve its objective impeccably while missing what its user sought. The converse is equally instructive: receiving a correct proof, perhaps one produced by an artificial system, need not bring understanding with it (Tsoukalas et al., 2026). Neither the feeling nor the result, taken alone, necessarily supplies the experience and achievement the person was pursuing.

An individualized model could distinguish these cases by taking account of what was understood, how the person arrived at it and what the achievement meant in their life. It might even predict their judgments more accurately than they could predict them beforehand. But the choice to treat a particular judgment as evidence of improvement is not a result that follows from a sufficiently detailed model of neuronal dynamics. A criterion based on what the subject now recognizes as desirable, or on what they wanted before the intervention, cannot be extracted from causal predictions alone. Nor will the earlier and later judgments always agree.

Where a person has encountered similar outcomes, their history may provide a good guide. For experiences unlike anything they have known, there may be no settled preference to consult. The relevant judgment may have to form in the encounter itself. And by then the person making it may no longer be quite the person who originally asked for the change.

## Changing the Judge

Neuroscaping can alter not only the experience being evaluated, but presumably also the processes by which it is evaluated. Learning, therapy and ordinary experience already do this; neuroscaping could make it more deliberate and extensive. A person who seeks relief from anxiety might later cease to value the caution that anxiety once supplied. Someone who wishes to feel less grief may come to attach less importance to the relationship whose loss occasioned it. In each case the intervention could change both the answer and the person answering.

Which judgment should count? The one made before the intervention, when its consequences were unfamiliar? Or the one made afterwards, by someone whose desires may have changed as a result? Neither answer will do in every case. Giving the earlier self an absolute veto would rule out transformations people deliberately seek. Giving the later self the last word could make almost any intervention appear successful. Someone who sought freedom from distress could emerge indifferent to things once dearly valued, and perfectly content with the change. That contentment is part of the outcome to be judged, not an independent audit of it.

The intervention can help determine which version of the person will answer the next question. Personal neuroscaping therefore cannot always be treated as the pursuit of a fixed preference waiting to be discovered. The subject who helps supply the criterion may change along with the experience being optimized (Paul, 2014). Predicting that change would make the disagreement visible; it would not resolve the authority of the competing judgments.

A model used to steer a person must be tested against the effects of steering. Writing can change the relationships on which a decoder was trained, so accurate prediction from passive observations does not establish how it will perform during intervention. Reading and writing test and refine one another. The person's responses also test something different: whether the selected criterion adequately expresses what they seek. Where preferences form through encounters with unfamiliar possibilities, specifying the destination and exploring it cannot be completed independently. Such transformations require a closed-loop component, with outcomes informing both the causal model and the criterion. These are distinct corrections: learning what an intervention does cannot by itself decide what the next intervention should pursue.

For such personal transformations, specifying a target and trying to realize it cannot be completed independently. A description leaves differences unresolved, and whether those differences matter has to be learned from the person's response. With unfamiliar experiences, even the terms of that assessment may emerge only in the encounter; the intervention can then alter the preferences brought to the next one. Predictions can inform the initial choice, but those responses must remain able to correct what the model treats as desirable. Under these conditions, neuroscaping seems to require at least some closed-loop component: outcomes inform both the next intervention and the criterion by which it is chosen, while disagreements between earlier and later evaluations should remain available for reconsideration. Neither earlier intentions nor later satisfaction can settle the value of the whole transformation. None of this authorizes continuing an intervention against the person's refusal. Feedback provides evidence for continuing to choose; it cannot certify that the choices are wise.

## Saints and Monsters

Larry Niven gave the prospect a human face in his wireheads, people who stimulated their brains directly, sometimes at the expense of everything else (Niven, 1969, 1980). The inducible reward need not be simple pleasure; it could be a convincing sense of fulfillment, unaccompanied by the activities, relationships or accomplishments that once made such fulfillment worth having.

Nor is the worst outcome necessarily withdrawal from the world. A person could seek patience, compassion or self-command and use the resulting capacities to help others. Another could cultivate immunity to remorse, a taste for domination, or the ability to deceive without discomfort. Still another might acquire intellectual capacities that dwarf our present accomplishments. Nothing in neuroscaping as a method favors the saint over the monster. Greater capability and greater worth are different things, and greater capability can serve either.

The problem becomes less private when such transformations affect others. Enhanced powers of persuasion could make difficult ideas easier to explain or make vulnerabilities easier to exploit. Disapproval of the resulting person's tastes does not settle the difference. The question concerns what they do with their powers, including how they affect people who never chose to be involved.

## Who Wrote the Objective?

An intervention does not translate its user's wishes directly into neuronal dynamics. A person asks to be less anxious, more capable of understanding, or simply happier; a controller works with neural measurements, behavioural reports and whatever perturbations its hardware can deliver.

Even then, somebody must decide how the person's words become a target the machine can recognize. Engineers choose which signals to measure, how to weight them, when to stop, and what counts as failure. A controller rewarded for reducing reported distress might actually relieve distress, or it might make the person less inclined to report it. A company may prefer an intervention that keeps people returning to one that leaves them satisfied and able to leave. Nothing about a closed loop ensures that its objective belongs to the person inside it. The choice of a criterion remains a moral judgment, as it does when ethical commitments become operational constraints in an Artificial Moral Core (Fekete, manuscript in preparation).

Olds and Milner (1954) showed that rats repeatedly pressed a lever to stimulate regions of their own brains. Later studies found animals neglecting food for stimulation during restricted daily sessions (Routtenberg & Lindy, 1965). With continuous access to both, however, self-deprivation disappeared (Frank et al., 1981). The lesson is not that direct reward inevitably destroys an animal's concern for survival. The arrangement of choices devised by the experimenter changed whether self-stimulation displaced feeding.

## Deeds, Not Minds

There is much that can be regulated without prescribing a person's preferred mental life. A stimulator that conceals adverse effects, an employer making enhancement a condition of employment, or a controller that reports success while failing to deliver the change promised presents problems of safety, consent and fidelity (Farah et al., 2004; Hamilton et al., 2011). A regulator can require evidence for the declared effect and an account of the resolution at which success has been assessed. Where the desired change is itself unsettled, that uncertainty must remain explicit. These are familiar obligations of those who build, sell or administer interventions. Neuroscaping changes what they need to establish, not the grounds on which honesty and authorization matter.

Regulating the desired mental endpoint is a different undertaking. Imagine a rule that permits mood interventions only insofar as they restore a prescribed range of emotional functioning. For some psychiatric conditions such a range may be a useful clinical reference. But why should it delimit the experiences available to an adult who can understand the proposed intervention and its uncertainties, weigh them against their purposes, and refuse it? A preference for tranquility over excitement, or for an unusual sensory life over an ordinary one, is not a malfunction merely because it falls outside a statistical norm. The authority supplying an official ideal of acceptable experience is deciding what sorts of minds its citizens may have.

The distinction I defend concerns conduct rather than mere potential. Exceptional powers of persuasion are not themselves manipulation, any more than physical strength is assault. Threats, preparations for harm and harmful uses of a capacity can warrant intervention; its mere possession cannot supply the same justification. Otherwise, almost any capacity could be restricted by imagining a sufficiently objectionable use. Selecting an approved human type through control of reproduction is eugenics (National Human Genome Research Institute, n.d.); imposing an approved mental life on existing people is a form of mind control. Greater power can increase what people owe others when they act. It does not make their inner lives public property.

The freedom to seek mental alteration, and to refuse it, has been defended under the rubric of cognitive liberty (Bublitz, 2013; Ienca & Andorno, 2017; Istace & Van Assche, 2026). That freedom rests on a familiar commitment to people's authority over their own lives, not on a discovery about the limits of a decoder (Mill, 1859, ch. IV). The modelling problem strengthens the objection to an imposed ideal: an external description can also miss what matters to the person. But even a perfect prediction of their preferences would not give its author ownership of the life predicted. Nor is the status quo made neutral by being familiar (Bostrom & Ord, 2006).

This does not make the use of that freedom ethically indifferent. People can be urged to cultivate courage, understanding or concern for others without being forced into an approved temperament. The distinction between encouragement and compulsion is familiar (Mill, 1859, ch. IV). Neuroscaping gives self-cultivation new means; it does not turn every opportunity to improve into a licence for someone else to command the improvement. An ethic of growth can guide what people make of themselves while leaving room to reject its advice.

## Beyond Our Measure

There is a further difficulty with prescribing what neuroscaping ought to make of people. The examples so far concern capacities humans already possess, however radically extended: memory, emotion, intelligence, self-control. Yet neuroscaping need not be confined to more or less of the familiar. It could eventually produce capacities and experiences beyond the range against which present judgments of flourishing have been tested. The problem would then be one of extrapolating evaluations into a part of the map no one has yet occupied.

The landscape need not be fixed by our present anatomy. Learning already involves structural change, including synaptic remodelling and the production of new myelinating cells (Xu et al., 2009; McKenzie et al., 2014), while stimulation can leave lasting changes in cortical responses (Jackson et al., 2006). A theory relating representational efficiency to brain organization could guide a more deliberate use of such processes. By drawing on mechanisms of growth and maturation, interventions might eventually extend processing capacity through altered connectivity, cellular differentiation or the integration of additional neurons. The contribution would have to be assessed through the dynamics of the developing system as a whole. More tissue alone would establish no improvement.

Such structural development would change which dynamics are possible and what subsequent interventions could achieve. If the new capacities also improved our ability to understand and direct further change, the person and the means of changing them could develop together. If the brain itself were enlarged, the body housing it might also need to change, for instance through a larger skull and stronger neck support. Engineered genetic circuits can already organize simple multicellular structures (Toda et al., 2018); much more extensive control of tissue development might eventually support neural elaboration. The point is that exploration could alter its own possibilities, as well as the abilities with which a person evaluates them.

Consider the authority of a toddler to pronounce on what constitutes a good adult life. There is much a child can appreciate about adults, yet a toddler is hardly in a position to judge whether the demands and pleasures of raising a family, conducting research or pursuing a difficult vocation make those lives worth living. The obstacle is not simply a shortage of information. The child lacks some of the capacities and experiences through which the information acquires its significance. An adult asked to prescribe the flourishing of a radically enhanced human might be in a similar position. The unfamiliarity of the experience is the difficulty, as in the problem of transformative choice (Paul, 2014).

Reports from those who undertake such transformations would help chart the new territory. But they would come from people whose criteria of judgment may themselves have changed. Their enthusiasm would be valuable evidence about lives outsiders could not yet understand, without settling whether every change that produced it was warranted. Experience gives access to what outsiders could not previously know; it does not confer authority over other people's lives, or guarantee an impartial verdict on one's own. The present generation has reason to govern the risks of the transition without supposing it can specify the destination in detail.

## The Price of Admission

An unequal distribution of neuroscaping could eventually give the enhanced greater influence over who may gain access next. This makes the familiar question of affordability unusually serious. What would be unequally distributed is not simply a benefit, but the capacity to acquire further benefits and to affect the rules under which everybody else competes (Farah et al., 2004; Greely et al., 2008).

Suppose improvements in learning, self-regulation or reasoning become available, initially at considerable expense. Those already enjoying material advantages would be first to acquire the new capacities. The capacities could then help them accumulate more resources, which would purchase still more effective neuroscaping. Education already participates in such feedback; neuroscaping could accelerate it and extend it to capacities that presently resist deliberate training. In the extreme, wealth would buy increasing capability, and capability would buy increasing wealth.

Even people who would rather remain as they are could be drawn in. If competitors can buy concentration, emotional resilience or extraordinary powers of persuasion, declining to do so has consequences in school, at work and in politics. A choice may be voluntary in law while becoming expensive in practice. The stakes are therefore not exhausted by whether a device is affordable: they include whether some people can afford not to change.

Nothing about the available technology establishes that this outcome is inevitable. Wider access could narrow gaps instead, and UNESCO's Recommendation on the Ethics of Neurotechnology (2025) addresses equitable access. Yet by changing the capacities with which people compete, neuroscaping could also affect who has the power to determine future access. A society has reasons to resist that prospect without first prescribing the kinds of minds its competent members ought to cultivate.

## Before the Choice

Children make the question of choosing for someone else unavoidable. A young child cannot judge changes that may shape the capacities and preferences through which their future life will be evaluated. Yet parents and educators already help make that future chooser. They select experiences, cultivate habits and open or close possibilities long before a child can assess the choices. Neuroscaping extends these powers without creating a new kind of parental authority. Feinberg's account of the child's right to an open future addresses a familiar obligation whose application becomes more demanding as the available means become more effective (Feinberg, 1980/1992).

Imagine parents choosing not merely to encourage perseverance but to make their child unusually compliant, fiercely competitive, or largely indifferent to social disapproval. Each choice could be defended as advantageous in some circumstances. But whose advantage, and according to whom? The distinction between developing a child's capacities and making the child convenient to others already matters in education. More direct intervention makes the consequences of crossing it potentially greater. Parents' intentions and the child's eventual approval cannot alone establish that their choice served the child's interests.

Present evidence gives substantial reasons for restraint. Maslen and colleagues (2014) discuss how enhancement involving developmental trade-offs can close future options and argue for greater caution than with therapy. Pediatric stimulation studies provide some safety evidence within studied protocols, but insufficient long-term evidence to evaluate ambitious interventions on personality and values (Salehinejad & Siniatchkin, 2024). The burden follows from what is being chosen for the child and how little is known about its consequences. It does not follow simply from the fact that the means are neural.

Nor can restoration remain the permanent boundary of what parents owe. Education develops capacities children never previously possessed; its justification does not depend on correcting a pathology. If an enhancement became a sufficiently safe, reliable and accessible means of developing capacities children need to lead lives of their own, providing it could become an obligation, perhaps one society could enforce as it does education (compare Mill, 1859, ch. V). Lasting benefits, developmental trade-offs and fair access would need to be established far beyond what presently supports such a requirement. The possibility follows from a familiar duty of care: new means can change what fulfilling that duty requires. It would not authorize a prescribed personality or set of ambitions.

## YES!

There are compelling reasons to develop means of relieving suffering and restoring capacities lost to disease or injury. Neuroscaping should unquestionably be put to such uses. But why stop at restoration? The average human mind is an outcome of evolution, not a standard of what a mind ought to be. If it becomes possible to improve our ability to understand, create, regulate our emotions, or even experience forms of beauty presently inaccessible to us, why should remaining as we are enjoy any special presumption in its favour?

The prospect is not morally innocent. Neuroscaping could make people happy with transformations their earlier selves would have abhorred, or make them more effective at harming others. Competition could also turn the freedom to enhance into a costly freedom to refuse. These are serious reasons to govern how neuroscaping is developed and used, and to hold people answerable for the consequences of their choices. But freedom that extends only to outcomes already known to be good is not much freedom. A real choice includes the possibility of being terribly wrong.

Artificial intelligence gives this prospect a further urgency. Systems are beginning to solve mathematical problems and generate literary and artistic work of a sophistication that would recently have seemed out of reach (Tsoukalas et al., 2026; Doshi & Hauser, 2024). These are welcome developments if the aim is to obtain a proof or a picture. But a mathematician does not struggle with a proof solely to possess its conclusion, any more than a painter paints merely to acquire another picture. Much of the beauty and purpose lie in the activity itself.[1]

Delegation is nevertheless tempting. It takes less effort to obtain a result than to acquire the ability to produce it, and the better our machines become, the easier it is to let them determine even what is worth attempting. AI need not make us less capable: an assistant designed to require independent thought can preserve or improve skills that a more obliging interface erodes (Noessel, 2026). But better access to another intelligence does not, by itself, extend our own. One can readily imagine becoming an increasingly sophisticated consumer of achievements one can neither produce nor fully appreciate.

As the capacities of our machines compound, learning to use ever more powerful crutches seems an increasingly modest ambition. If enlarging our ability to learn and explore eventually helps us devise still better ways to enlarge those abilities, human development could itself become cumulative, perhaps even exponential. This lies far beyond what contemporary interventions can deliver. Yet it may be the only prospect that leaves us neither increasingly at the mercy of powers we cannot understand nor obliged to arrest their development to preserve a place for ourselves. The obligation is to pursue that possibility and make it accessible, not to require anybody to use it. Indeed, the competitive pressures created by enhancement are precisely why keeping the choice meaningful will be difficult.

To the question in the title, my answer is yes. Some will use that freedom to make themselves better, others to become much worse, and there will be cases no present standard can confidently classify. Yet resigning ourselves to the cognitive and experiential limits with which evolution furnished us, while the powers we have brought into the world continue to grow, seems a rather poor idea of progress. What more could one want from neuroscaping than the possibility of becoming capable of things one cannot yet even imagine wanting?

---

[1] I have argued elsewhere that digital computers do not instantiate the dynamics necessary for phenomenal experience (Fekete & Edelman, 2012). If so, a result produced by a machine and understood by no person may be a discovery whose insight has been experienced by no one. Different machines might one day be sentient, in which case they would deserve the moral consideration due to any creature that experiences.

## Competing interests

The author is Chief Science Officer at brain.space, a company developing technology for measuring and modelling human neural and physiological activity. The company does not currently offer neural-stimulation devices.